\documentclass{article}
\usepackage{spconf,amsmath,graphicx}
\usepackage{url}
\usepackage{algorithm,algorithmic}
\usepackage{bm}

\usepackage{enumitem}
\usepackage{amsfonts}
\setlist{nosep, leftmargin=14pt}    

\usepackage{mwe} 
\usepackage{multirow}

\title{Deep Clustering Survival Machines with Interpretable Expert Distributions}
\name{Bojian Hou$^{\star}$, Hongming Li$^{\star}$, Zhicheng Jiao$^{\dagger}$, Zhen Zhou$^{\star}$, Hao Zheng$^{\star}$, Yong Fan$^{\star}$}

\address{$^{\star}$ Department of Radiology, Perelman School of Medicine, University of Pennsylvania, USA \\
    $^{\dagger}$ Department of Diagnostic Imaging, Warren Albert Medical School, Brown University, USA} 

\begin{document}
%
\maketitle
\begin{abstract}
Conventional survival analysis methods are typically ineffective to characterize heterogeneity in the population while such information can be used to assist predictive modeling. In this study, we propose a hybrid survival analysis method, referred to as deep clustering survival machines, that combines the {\it discriminative} and {\it generative} mechanisms. Similar to the mixture models, we assume that the timing information of survival data is {\it generatively} described by a mixture of certain numbers of parametric distributions, i.e., {\it expert distributions}. We learn weights of the expert distributions for individual instances according to their features {\it discriminatively} such that each instance's survival information can be characterized by a weighted combination of the learned constant expert distributions. This method also facilitates interpretable subgrouping/clustering of all instances according to their associated expert distributions. Extensive experiments on both real and synthetic datasets have demonstrated that the method is capable of obtaining promising clustering results and competitive time-to-event predicting performance. Code is available at \url{https://github.com/BojianHou/DCSM}.
\end{abstract}
\section{Introduction}
\label{sec:intro}

In survival analysis, it is desired to know individual subjects' probability of an event of interest to occur such as the occurrence of a disease or even death beyond a certain time $t$ according to their data \slash features $X$ \cite{flynn2012survival}. As a result, the probability can be modeled as a survival function $S(\cdot|X)=P(T>t|X)$. This task is also referred to as {\it time-to-event prediction}, and one of its main challenges is {\it censoring} when the event outcomes of some instances are unobservable after a certain time or some instances do not experience any event during follow-up. 

Many methods have been proposed for time-to-event prediction in survival analysis. The most conventional and prevalent method is a semi-parametric method called the Cox PH model \cite{cox1972regression}. It assumes that the hazard rate for every instance is constant over time known as the proportional hazard (PH) assumption. 
Some nonparametric methods such as Kaplan-Meier \cite{bland1998survival}, Nelson-Aalen \cite{klein1991small}, and Life-Table \cite{tarone1975tests} are also widely used in survival analysis. Nevertheless, they suffer from the curse of dimensionality. Survival analysis also attracts the attention of the machine learning community and many machine learning methods \cite{ranganath2016deep,katzman2018deepsurv,kvamme2019time,wang2019machine,lin2021empirical,jiao2022integration} have been developed to reveal the relationship between the features and the survival information. Particularly, a fully parametric method, referred to as deep survival machines (DSM) \cite{nagpal2021deep}, has demonstrated competitive predicting performance compared with state-of-the-art methods. Nevertheless, DSM learns different base distributions for different instances, which makes its inner mechanism hard to interpret~\cite{hou2018learning,hou2020learning}.

In addition to the time-to-event prediction task, the task of clustering cohorts is also crucial in survival analysis. With the clustering results, clinicians can provide customized treatments~\cite{hou2021clinical} for groups with different risks. The methods mentioned above usually cluster the cohorts in a post-hoc way, i.e., they will artificially stratify the cohorts according to the predicted risks. This usually leads to even groups, thus lacking interpretability. Lately, two studies consider both clustering and time-to-event prediction simultaneously and they can cluster data in an uneven manner. Particularly, survival clustering analysis (SCA) \cite{chapfuwa2020survival} assumes that the latent space is a mixture of distributions and uses the truncated Dirichlet process to realize the automatic identification of the cluster numbers. However, SCA cannot control the number of clusters and thus cannot validate its advantages compared to those post-hoc methods. Variational deep survival clustering (VaDeSC) \cite{manduchi2021deep}, as a fully generative method, uses Gaussian mixture distribution to model the features in a latent space and uses the Weibull distribution to model the survival timing information. This work builds a good bridge between the features and survival information by jointly optimizing both likelihoods. However, there is a trade-off between the discriminative and generative learning paradigms. A fully generative framework may not be a good fit for all kinds of data since it is hard to let both the features and survival information obey the prior assumption of distributions at the same time.

\begin{figure*}[htb]
\begin{minipage}[b]{1.0\linewidth}
  \centering
  \centerline{\includegraphics[width=14.3cm]{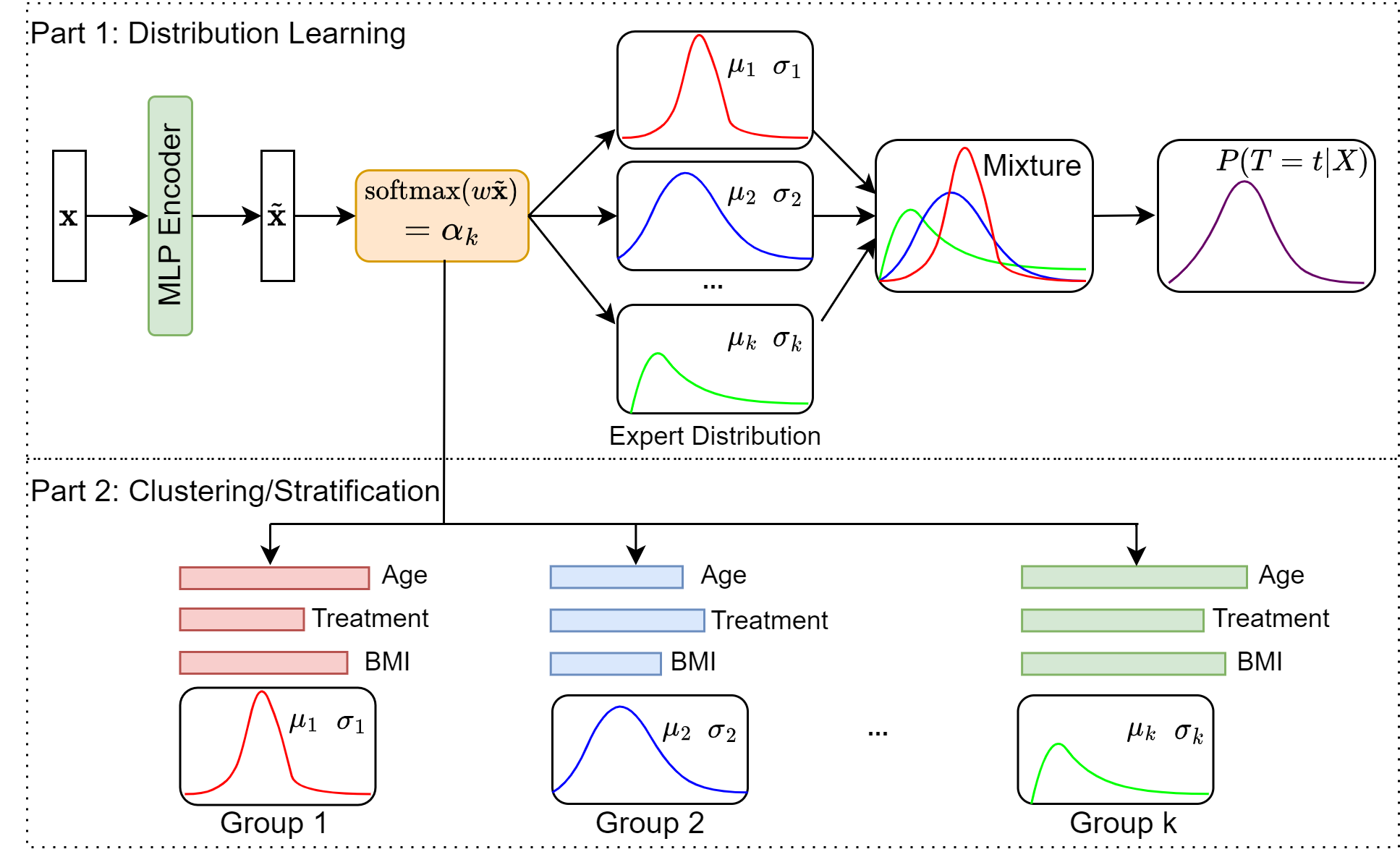}}
\end{minipage}
\caption{The model structure of the proposed DCSM. Part~1 learns each instance's survival function by a weighted combination of the expert distributions. Part~2 clusters instances by the learned weights allocated to each expert distribution.}
\label{fig:model}
\end{figure*}

In this study, we propose a hybrid method to leverage both the discriminative and generative strategies. Specifically, we assume that there are certain numbers of expert distributions in a latent space and each expert distribution can be modeled by parameterized Weibull distributions in a generative way. The survival function for each instance is a weighted combination of all the expert distributions and the weight for each instance is learned by a multi-layer perceptron (MLP) directly from the features in a discriminative manner. Consequently, we can naturally cluster all the instances according to their weights allocated to different expert distributions. In summary, our contributions are threefold:
\begin{itemize}
    \item We propose a hybrid survival analysis method that integrates the advantage of discriminative and generative ideas, and can perform both clustering and time-to-event prediction simultaneously.
    \item We conduct extensive experiments on several real-world datasets and abundant synthetic datasets, and the results show promising clustering results as well as competitive time-to-event prediction performance. 
    \item Our method is interpretable in that the expert distributions are constant for all the instances. Different weight shows different attention to the expert distributions and thus we can easily tell which subgroup the instance belongs to.
\end{itemize}

\section{Methods}
\label{sec:method}
The data we tackled are right-censored, i.e., our data $\mathcal{D}$ is a set of tuples $\{\textbf{x}_i,t_i,\delta_i\}_{i=1}^N$ where $\textbf{x}_i$ is the feature vector associated with the $i$th instance, $t_i$ is the last-followed time, $\delta_i$ is the event indicator, and $N$ is the number of instances. When $\delta_i=1$ (it means the $i$th instance is uncensored), $t_i$ will be the time when the event happens whereas when $\delta_i=0$ (it means the $i$th instance is censored), $t_i$ will be the time when the instance quits the study or the study ends. Denote the $\mathcal{D}_U$ as the uncensored subset where the corresponding event indicator $\delta=1$ and $\mathcal{D}_C$ as the censored subset where $\delta=0$.

In Part~1 of Fig.~\ref{fig:model}, the deep clustering survival machines are designed to learn a conditional distribution $P(T|X=\textbf{x})$ by optimizing the maximum likelihood estimation (MLE) of the time $T$. Similar to the mixture model learning paradigm, the conditional distribution $P(T|X=\textbf{x})$ is characterized by learning a mixture over $K$ well-defined parametric distributions, referred to as {\it expert distributions}. In order to use gradient-based methods to optimize MLE, we choose the Weibull distributions as the expert distributions that are flexible to fit various distributions and have closed-form solutions for the PDF and CDF: 
\[
    \text{PDF}(t)=\frac{\mu}{\sigma}\left(\frac{t}{\sigma}\right)^{\mu-1}e^{-\left(\frac{t}{\sigma}\right)^\mu}, \
    \text{CDF}(t)=e^{-\left(\frac{t}{\sigma}\right)^\mu},
\]
where $\mu$ and $\sigma$ are the shape and scale parameters separately.

In Part 1 of Fig.~\ref{fig:model}, the deep clustering survival machines (DCSM) is designed to learn a conditional distribution $P(T|X=\textbf{x})$ by optimizing the maximum likelihood estimation (MLE) of the time $T$. Similar to the mixture model learning paradigm, the conditional distribution $P(T|X=\textbf{x})$ is characterized by learning a mixture over $K$ well-defined parametric distributions, referred to as {\it expert distributions}. In order to use gradient-based methods to optimize MLE, we choose the Weibull distributions as the expert distributions that are flexible to fit various distributions and have closed-form solutions for the PDF and CDF: 
\begin{equation}
    \text{PDF}(t)=\frac{\mu}{\sigma}\left(\frac{t}{\sigma}\right)^{\mu-1}e^{-\left(\frac{t}{\sigma}\right)^\mu}, \
    \text{CDF}(t)=e^{-\left(\frac{t}{\sigma}\right)^\mu},
\end{equation}
where $\mu$ and $\sigma$ are the shape and scale parameters separately.

Part 1 of Fig.~\ref{fig:model} indicates that we firstly need to learn an encoder for the input features $\textbf{x}\in\mathbb{R}^d$ to obtain a compact representation $\tilde{\textbf{x}}\in\mathbb{R}^{d'}$. Here we use a multi-layer perceptron (MLP) $\phi_{\bm{\theta}}(\cdot)$ parameterized by ${\bm{\theta}}$ as the backbone model. This representation will be multiplied by a parameter $\bm{w}\in\mathbb{R}^{d'\times K}$ with {\it softmax} to obtain the mixture weight $\alpha_k,\ k=1,\ldots,K$ with respect to each ($k$th) expert distribution that is parameterized by $\mu_k$ and $\sigma_k$. The final survival distribution for the time $T$ conditioned on each instance is a weighted combination over all $K$ constant expert distributions. Eventually, we have a set of parameters $\Theta=\{{\bm{\theta}},\bm{w},\{\mu_k,\sigma_k\}_{k=1}^K\}$ to learn during the training process. Because $\mu_k$ and $\sigma_k$ are the same for different input instances, we can cluster each instance/subject according to the weight $\alpha_k$ that is allocated to each expert distribution, as illustrated in Part~2 of Fig.~\ref{fig:model}. Specifically, we assign an subgroup/cluster indicator $k$ to each instance when the instance’s corresponding weight $\alpha_k$ is the largest among all $K$ weights.

According to the framework of MLE, our goal is to maximize the likelihood with respect to the timing information $T$ conditioned on $\textbf{x}$. Given that the likelihood functions are different for uncensored and censored data, we calculate them separately. For the uncensored data, the log-likelihood of $T$ is computed as follows, where \textbf{ELBO} is the lower bound of the likelihood derived by Jensen’s Inequality:

\begin{algorithm}[t] 
  \caption{DCSM Training, Time-to-event Prediction and Clustering}
  \label{alg:DCSM alg}
  \begin{algorithmic}[1]
    \STATE \textbf{Input}: Dataset $\mathcal{D}$ consists of tuples $\{\textbf{x}_i,t_i,\delta_i\}_{i=1}^N$
    \STATE \textbf{Output}: Trained model $f=\{\phi_{\bm{\theta}},\bm{w},\{\mu_k,\sigma_k\}_{k=1}^K\}$, the estimated risk $r_i$ and the cluster label $k$ for each subject $i$
    \STATE Split $\mathcal{D}$ into a training set $\mathcal{D}_\text{tr}$ and a testing set $\mathcal{D}_\text{te}$
    \STATE \# {\it \textbf{Training Phase...}}
    \STATE Initialize a prior-model $f_\text{prior}=\{\phi_{\bm{\theta}_\text{prior}},\bm{w}_\text{prior},\mu,\sigma\}$
    \STATE Pre-train the model $f_\text{prior}$ with only one expert distribution parameterized by $\mu$ and $\sigma$ by maximizing (\ref{eq:loss_uncensored})+(\ref{eq:loss_censored}) on $\mathcal{D}_\text{tr}$ where the trained prior-model $f_\text{prior}$ will be used in (\ref{eq:prior})
    \STATE Initialize a formal model $f=\{\phi_{\bm{\theta}},\bm{w},\{\mu_k,\sigma_k\}_{k=1}^K\}$
    \STATE Train the model $f$ with multiple expert distributions parameterized by $\mu_k$ and $\sigma_k$ by minimizing (\ref{eq:all}) on $\mathcal{D}_\text{tr}$ and obtain the trained model $f$
    \STATE \# {\it \textbf{Time-to-event Prediction Phase...}}
    \STATE Calculate the weights $\{\alpha_{k}\}_{k=1}^K$ of all $K$ expert distributions for the $i$th subject by (\ref{eq:softmax})
    \STATE Obtain the risk $r_i$ of the $i$th subject by (\ref{eq:estimate risk})
    \STATE \# {\it \textbf{Clustering Phase...}}
    \STATE Obtain the cluster label $k$ for the $i$th subject based on the largest $\alpha_{k}$ 
  \end{algorithmic} 
\end{algorithm}

\begin{align}\label{eq:loss_uncensored}
    &\ln\mathbb{P}(\mathcal{D}_U|\Theta)=\ln\left(\Pi_{i=1}^{|\mathcal{D}_U|}\mathbb{P}(T=t_i|X=\textbf{x}_i,\Theta)\right) \nonumber\\
    &=\sum\nolimits_{i=1}^{|\mathcal{D}_U|}\ln\left(\sum\nolimits_{k=1}^K\mathbb{P}(T=t_i|\alpha_{k},\mu_k,\sigma_k)\mathbb{P}(\alpha_{k}|X=\textbf{x}_i,\bm{w})\right) \nonumber\\
    &=\sum\nolimits_{i=1}^{|\mathcal{D}_U|}\ln\left(\mathbb{E}_{\alpha_{k}}\sim(\cdot|\textbf{x}_i,\bm{w})[\mathbb{P}(T=t_i|\alpha_{k},\mu_k,\sigma_k)]\right) \\
    &\geq\sum\nolimits_{i=1}^{|\mathcal{D}_U|}\left(\mathbb{E}_{\alpha_{k}}\sim(\cdot|\textbf{x}_i,\bm{w})[\ln\mathbb{P}(T=t_i|\alpha_{k},\mu_k,\sigma_k)]\right) \nonumber\\
    &=\sum\nolimits_{i=1}^{|\mathcal{D}_U|}\left(\text{softmax}_K(\ln\text{PDF}(t_i|\mu_{k},\sigma_{k}))\right)=\textbf{ELBO}_U(\Theta). \nonumber
\end{align}

Similarly, the log-likelihood of $T$ for the censored data is:
\begin{equation}\label{eq:loss_censored}
\begin{split}
    &\ln\mathbb{P}(\mathcal{D}_C|\Theta)=\ln\left(\Pi_{i=1}^{|\mathcal{D}_C|}\mathbb{P}(T>t_i|X=\textbf{x}_i,\Theta)\right) \\
    &\geq\sum\nolimits_{i=1}^{|\mathcal{D}_C|}\left(\mathbb{E}_{\alpha_{k}}\sim(\cdot|\textbf{x}_i,\bm{w})[\ln\mathbb{P}(T>t_i|\alpha_{k},\mu_k,\sigma_k)]\right) \\
    &=\sum\nolimits_{i=1}^{|\mathcal{D}_C|}\left(\text{softmax}_K(\ln\text{CDF}(t_i|\mu_{k},\sigma_{k}))\right)=\textbf{ELBO}_C(\Theta).
\end{split}
\end{equation}
In addition, to stabilize the performance, we incorporate prior knowledge for $\mu_k$ and $\sigma_k$. 
Specifically, we minimize the prior loss $L_{prior}$ to make them as close as possible to the $\mu$ and $\sigma$ from the prior model:
\begin{equation}\label{eq:prior}
L_{prior}=\sum\nolimits_{k=1}^K\|\mu_k-\mu\|_2^2+\|\sigma_k-\sigma\|_2^2.
\end{equation}
where the prior model is learned by the same MLE framework with a single expert distribution that is still Weill distribution. The final objective $L_{all}$ is the sum of the negative of the log-likelihoods of both the uncensored and censored data in addition to the prior loss where $\lambda$ is a trade-off hyperparameter:
\begin{equation}\label{eq:all}
    L_{all}=L_{prior}-\textbf{ELBO}_U(\Theta)-\lambda\cdot\textbf{ELBO}_C(\Theta).
\end{equation}

The implementing details are as follows. First, we split the dataset $\mathcal{D}$ into a training set $\mathcal{D}_\text{tr}$ and a testing set $\mathcal{D}_\text{te}$. In the training phase, we first initialize a prior-model $f_\text{prior}=\{\phi_{\bm{\theta}_\text{prior}},\bm{w}_\text{prior},\mu,\sigma\}$ where the prior-model only contains one expert distribution parameterized by $\mu$ and $\sigma$. In our implementation, we use PyTorch~\cite{NEURIPS2019_9015} to conduct the model initialization. Then we pre-train the prior-model $f_\text{prior}$ by maximizing the likelihood (\ref{eq:loss_uncensored})+(\ref{eq:loss_censored}) on $\mathcal{D}_\text{tr}$. In this way, the learned $\mu$ and $\sigma$ from the prior model can be used in (\ref{eq:prior}). Then we initialize the formal model $f=\{\phi_{\bm{\theta}},\bm{w},\{\mu_k,\sigma_k\}_{k=1}^K\}$ and train it by minimizing (\ref{eq:all}) on $\mathcal{D}_\text{tr}$. After we obtain the trained model $f$, we can conduct time-to-event prediction and clustering simultaneously. To do that, we first need to calculate the weights $\{\alpha_{k}\}_{k=1}^K$ of all $K$ expert distribution for the $i$th subject by the {\it softmax} on $\bm{w}^\top\phi_{\bm{\theta}}(\textbf{x}_i)$:
\begin{equation}\label{eq:softmax}
    \alpha_{k}=\frac{\exp((\bm{w}^\top\phi_{\bm{\theta}}(\textbf{x}_i))_k)}{\sum_{j=1}^K\exp((\bm{w}^\top\phi_{\bm{\theta}}(\textbf{x}_i))_j)}
\end{equation}
For time-to-event prediction, we use the weights to conduct weighted combination for all the CDF value given a specific time $t$ which is the time horizon $t_{max}$ in our case. Then the risk for the $i$th subject $r_i$ is estimated by
\begin{equation}\label{eq:estimate risk}
\begin{split}
    r_i&=1-\sum_{k=1}^K\mathbb{P}(T\leq t_{max}|\alpha_{k},\mu_k,\sigma_k) \\
    &=1-\sum_{k=1}^K\alpha_{k}\text{CDF}(t_{max})\\
    &=1-\sum_{k=1}^K\alpha_{k}\exp\left(-\left(\frac{t_{max}}{\sigma_k}\right)^{\mu_{k}}\right).
\end{split}
\end{equation}
For clustering, we just assign the index $k$ to the $i$th subject if $\alpha_{k}$ is the largest among all the $K$ weights. The algorithm is summarized in Algorithm~\ref{alg:DCSM alg}.

\begin{table*}[t]
    \centering
    \caption{C Index and LogRank results compared to Cox PH, Deep Cox, DSM, SCA, and VaDeSC. The best ones are bold.}
    \begin{tabular}{c|c|c|c|c|c}
    \hline
Metric & Dataset	& SUPPORT &	PBC	& FRAMINGHAM &	FLCHAIN\\
\hline
\multirow{6}{*}{C Index} & Cox PH &	0.8401±0.0070 &	\textbf{0.8476±0.0126} &	0.7580±0.0063 &	0.7984±0.0046 \\
 & Deep Cox &	0.8053±0.0058 &	0.8474±0.0181 &	 \textbf{0.7612±0.0057} &	0.7893±0.0063 \\
 & DSM	& 0.8300±0.0045 & 	0.8363±0.0133 & 	0.7593±0.0050 &	\textbf{0.8009±0.0036} \\
 & SCA &	0.8203±0.0121 &	0.8251±0.0258 & 0.5311±0.1235 &	0.7467±0.0091\\
 & VaDeSC & \textbf{0.8419±0.0041} & 0.8278±0.0085 & 	0.5802±0.0406 &	0.7886±0.0100 \\
 & DCSM (Ours) &	0.8305±0.0028 &	0.8359±0.0109	& 0.7530±0.0053 &	0.7916±0.0074\\
      \hline
\multirow{6}{*}{LogRank} & Cox PH  &	500.3282±60.4977  &	198.2686±17.3940  &	576.1450±22.9621  &	399.0243±25.7657\\
& Deep Cox  &	326.1931±54.7026  &	203.3091±22.8343  &	593.7317±14.4697  &	403.4643±35.8034\\
& DSM	 & 563.4841±0.0045  &	196.0912±0.0133	 & 587.5718±0.0050	 & 406.4549±0.0036\\
& SCA	 & 212.5712±26.2629	 & 260.5682±67.4875  &	278.3525±51.1866  &	536.1056±109.1680\\
& VaDeSC  &	196.8495±19.6887  &	118.9605±77.4716  &	348.5500±697.1000  &	95.5291±108.9488\\
& DCSM (Ours)	 & \textbf{1067.6184±271.6551}	 & \textbf{302.5395±30.1043}  &	\textbf{751.9770±48.9725}  &	\textbf{571.0441±99.0101}\\
\hline
    \end{tabular}
    \label{tab:results of real data}
\end{table*}

\section{Experiments}
\label{sec:experiments}
We conducted extensive experiments to validate the effectiveness of the proposed method in terms of both time-to-event prediction and clustering. 

\begin{table}[h!]
    \centering
    \setlength{\tabcolsep}{4pt}
    \caption{Statistics of datasets used in the experiments. The time range $t_{max}$ in PBC is noted in years while others are noted in days. “FRAM” refers to “FRAMINGHAM”.} %
    \begin{tabular}{l|c|c|c|c}
    \hline
      Dataset & SUPPORT & PBC & FRAM &FLCHAIN  \\
      \hline
      Events (\%) & 68.11 & 37.28 & 30.33 & 30.07 \\
      \hline
      $N$ & 9105&	1945& 11627& 6524\\
      \hline
      $d$ (categorical)& 44 (26)& 25 (17)& 18 (10)& 8 (2) \\
      \hline
      $t_{max}$& 2029& 14.31& 8766& 5167\\
      \hline
    \end{tabular}
    \label{tab:Statistics of datasets}
    \vspace{-0.2cm}
\end{table}

\subsection{Datasets}
We conducted experiments on 4 real-world datasets (as shown in Table~\ref{tab:Statistics of datasets}) and 36 synthetic datasets with different numbers of instances, ranging in 200, 500, 1000, 3000, 5000, and 10000, and different numbers of features ranging in 10, 20, 50, 200, 500, and 1000. For all the synthetic datasets, the percentage of censoring was set to 30\%. The simulation process followed VaDeSC \cite{manduchi2021deep} except that we changed the distribution of the features from Gaussian to Uniform to validate the limitation of the fully generative method, which is discussed in Section~\ref{sec:results of synthetic}.  

\subsection{Baseline Methods, Metrics and Settings}
We compared our method to five methods. Two of them are the state-of-the-art methods SCA \cite{chapfuwa2020survival} and VaDeSC \cite{manduchi2021deep} which can perform both time-to-event prediction and clustering. The other three methods are Cox PH \cite{cox1972regression}, Deep Cox \cite{katzman2018deepsurv}, and DSM \cite{nagpal2021deep}, which only provide the time-to-event prediction function. We used their predicted risks to cluster data evenly.

We used two metrics to evaluate the performance of all the methods. Specifically, “concordance index” (C Index) was used to evaluate the time-to-event prediction performance. For the clustering task, we leveraged LogRank test to evaluate the performance. 

We conducted five-fold cross validation to estimate the C Index and LogRank measures and obtained their average values along with the standard deviation. The parameters were chosen by grid search. Specifically, the trade-off parameter $\lambda$ was chosen from [0.5, 0.75, 1], and the learning parameter step size was chosen from [1e-3, 1e-4]. The layer setting of the multiple perceptron was chosen from [[50], [50, 50]] where “50” is the number of neurons in each layer.

\begin{figure*}[t]
\begin{minipage}{0.24\linewidth}
  \centering
  \centerline{\includegraphics[width=\textwidth]{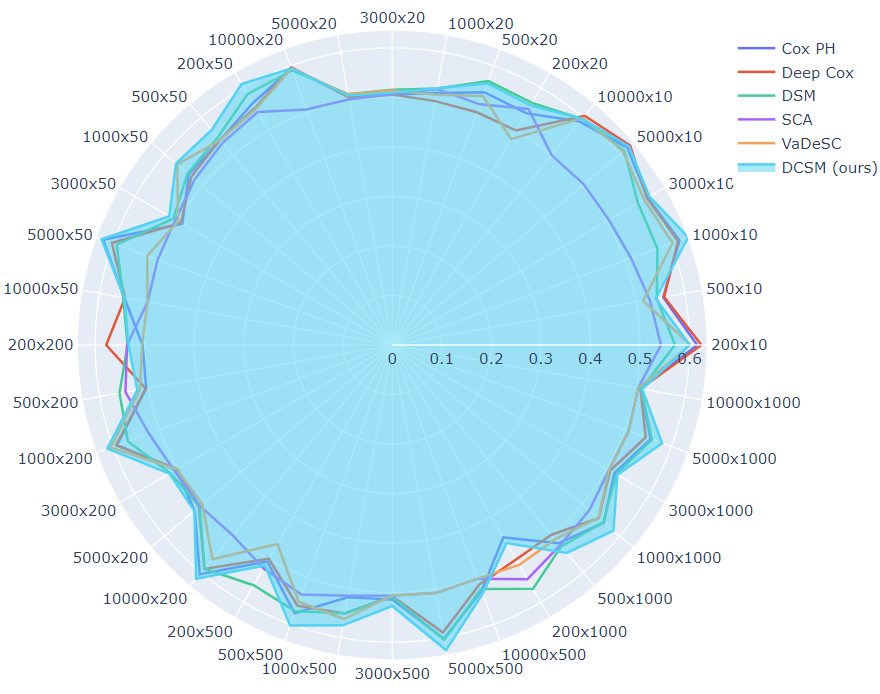}}
  \centerline{(a) C Index on synthetic data}\medskip
\end{minipage}
\begin{minipage}{0.24\linewidth}
  \centering
  \centerline{\includegraphics[width=\textwidth]{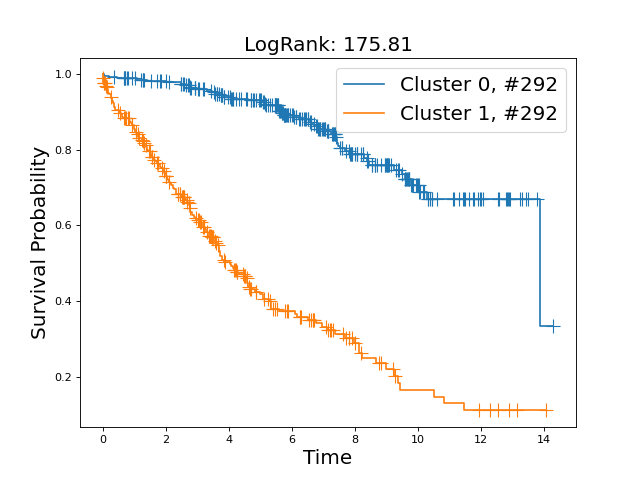}}
  \centerline{(b) KM plot of Cox PH}\medskip
\end{minipage}
\begin{minipage}{0.24\linewidth}
  \centering
  \centerline{\includegraphics[width=\textwidth]{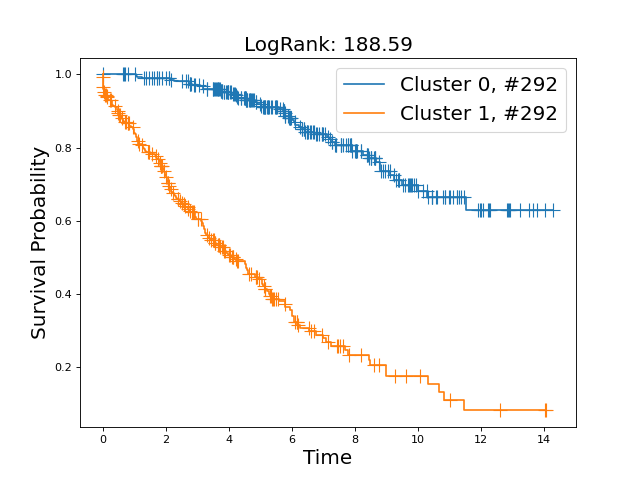}}
  \centerline{(c) KM plot of Deep Cox}\medskip
\end{minipage}
\begin{minipage}{0.24\linewidth}
  \centering
  \centerline{\includegraphics[width=\textwidth]{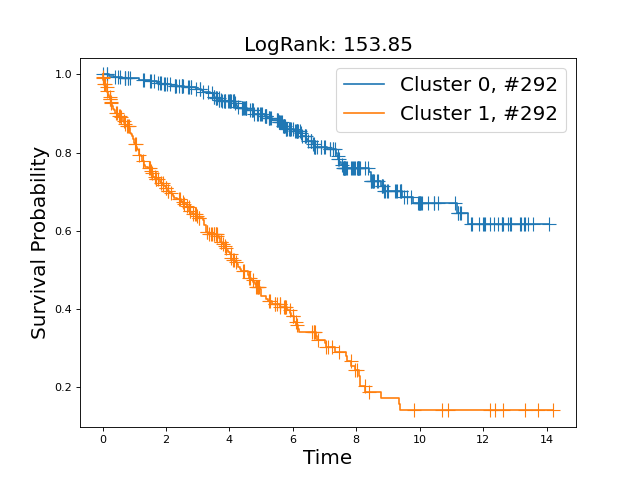}}
  \centerline{(d) KM plot of DSM}\medskip
\end{minipage}

\begin{minipage}{0.24\linewidth}
  \centering
  \centerline{\includegraphics[width=\textwidth]{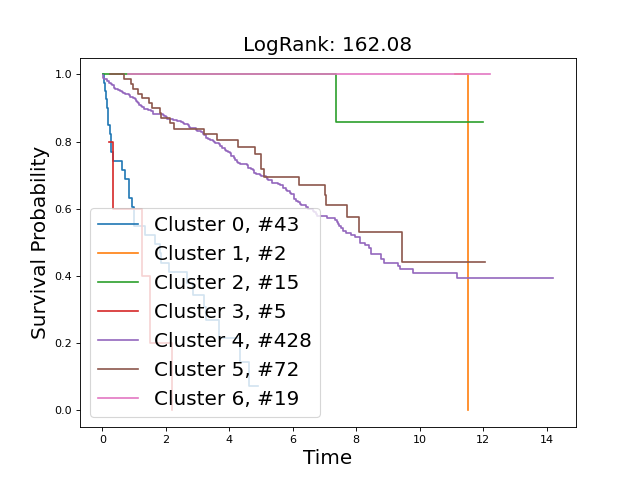}}
  \centerline{(e) KM plot of SCA}\medskip
\end{minipage}
\begin{minipage}{0.24\linewidth}
  \centering
  \centerline{\includegraphics[width=\textwidth]{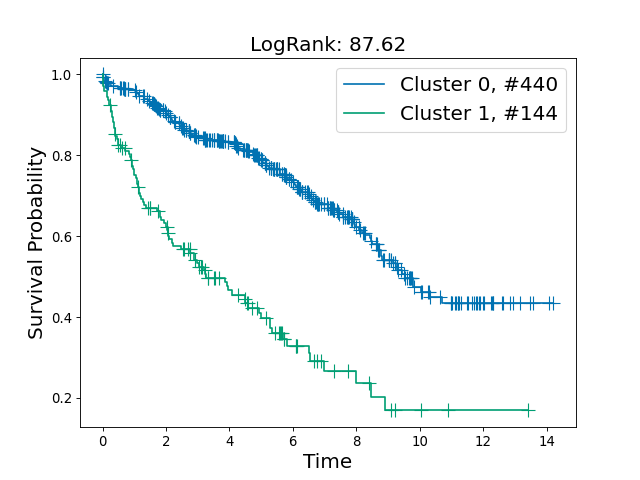}}
  \centerline{(f) KM plot of VaDeSC}\medskip
\end{minipage}
\begin{minipage}{0.24\linewidth}
  \centering
  \centerline{\includegraphics[width=\textwidth]{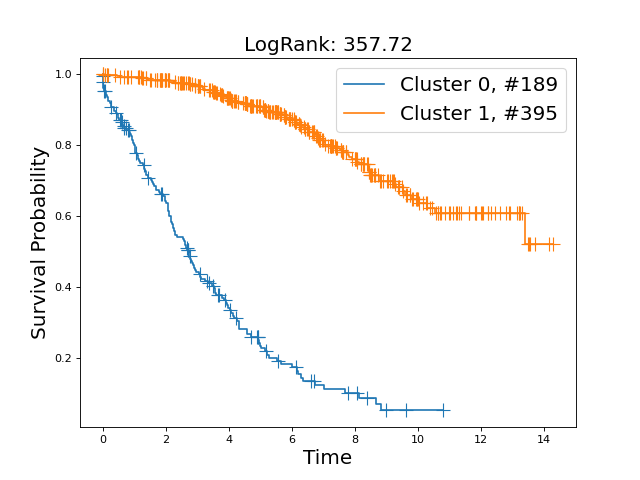}}
  \centerline{(g) KM plot of DCSM (ours) }\medskip
\end{minipage}
\begin{minipage}{0.24\linewidth}
  \centering
  \centerline{\includegraphics[width=\textwidth]{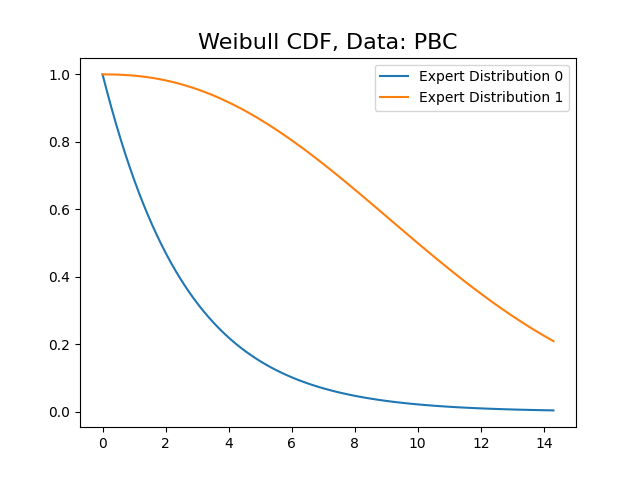}}
  \centerline{(h) Expert distribution of DCSM}\medskip
\end{minipage}
\caption{(a) The C Index comparison among the 36 synthetic datasets. A radar plot is used to illustrate the performance comparison. A bigger area means better performance. We fill the area of our method and we can see that on most synthetic datasets (30 among 36), the baseline methods’ curves fall inside our method. (b-g) The Kaplan-Meier plots of all the methods on data PBC. The cross mark means censoring. The learned expert distributions are shown in (h). The shape of the two expert distributions resembles our Kaplan-Meier curves, facilitating effective data stratification.}
\label{fig:res}
\end{figure*}

\subsection{Quantitative Results on Real Data}
Table~\ref{tab:results of real data} shows the C Index values on real data, including the average results of five independent runs and their standard deviations. These results indicated that our method achieved a competitive performance compared to other baselines. Although our model’s performance was not the best on some datasets, the difference with the best performance was not significant 
at a 95\% confidence interval.

Table~\ref{tab:results of real data} also summarizes the results of the LogRank tests. LogRank statistic evaluates how well the clustering results are regarding the survival information and 
with a larger value indicating a better performance. The results demonstrated that our method outperformed all the baselines. This could be more useful than the time-to-event prediction because such information can facilitate personalized treatment planning.

\subsection{Quantitative Results on Synthetic Data}
\label{sec:results of synthetic}
Fig.~\ref{fig:res}(a) shows the comparison of C Index values on synthetic data. We used radar plot to highlight the performance difference. The bigger the area surrounded by the curves, the better the performance. Fig.~\ref{fig:res}(a) demonstrated that our method generated the biggest area surrounded by the curve, indicating that our method outperformed all the baselines on 30 among 36 datasets. 
Our method learned the survival information generatively by assuming the survival information follows the Weibull distribution. As Weibull distribution is rather flexible and can simulate many different distributions from reality, therefore our method can fit well to the survival information and obtain the best performance in most cases.

VaDeSC as a fully parametric method also assumes the survival information obeys the Weibull distribution, but it assumes the features follow the Gaussian distribution whereas we generate the features using Uniform distribution. In this way, VaDeSC cannot model the feature distribution well and thus has inferior performance. Our method learns the features in a discriminative way. Thus our method can learn a likely pattern no matter what the real distribution of the features is.


\subsection{Qualitative Results on Real Data}
Kaplan-Meier (KM) curves according to the clustering results of all the methods are shown in Fig.~\ref{fig:res}(b-g). Due to the page limit, we only show the results on the PBC dataset. The LogRank of one trial of these methods was 175.81 (Cox PH), 188.59 (Deep Cox), 153.85 (DSM), 162.08 (SCA), 87.62 (VaDeSC), and 357.72 (DCSM, ours). It is worth noting that 
SCA and VaDeSC in Fig.~\ref{fig:res}(e, f) can automatically determine the numbers of instances in different groups. 
VaDeSC had more unbalanced results, which results in a low LogRank. Our method obtained the best performance. 
 Fig.~\ref{fig:res}(h) shows that the shapes of the two expert distributions resemble the KM curves, facilitating effective data stratification.

\section{Conclusion}
We propose a deep hybrid method that integrates the discriminative and generative strategies into one framework. Assuming the survival function for each instance is a weighted combination of constant expert distributions, our method is capable of learning the weight for each expert distribution discriminatively and the distribution of the survival information generatively. Extensive experimental results along with the quantitative and qualitative analyses have demonstrated the advantages of our method. The constant expert distributions also enhance the interpretability of data stratification.


\section{Compliance with ethical standards}
\label{ethics}
Our method complies with ethical standards. All the datasets we studied are public benchmark datasets. 


\bibliographystyle{IEEEbib}
\bibliography{refs}

\end{document}